\pdfoutput=1
\documentclass[11pt,a4paper]{article}
\usepackage[hyperref]{acl2021}
\usepackage{times}
\usepackage{float}
\usepackage{latexsym}
\usepackage{amsmath}
\usepackage{algorithmic}
\usepackage{algorithm}
\usepackage{appendix}
\usepackage{booktabs}
\usepackage{graphicx}
\usepackage{subcaption}

\usepackage{microtype}

\aclfinalcopy 

\title{Generating Personalized Dialogue via Multi-Task Meta-Learning}

\author{Lee Jing Yang\textsuperscript{1}, Lee Kong Aik\textsuperscript{2}, Gan Woon Seng\textsuperscript{3}\\
  School of Electrical and Electronic Engineering, Nanyang Technological University\textsuperscript{1,3} \\
  Institute for Infocomm Research, A*STAR\textsuperscript{2}\\
  \texttt{jingyang001@e.ntu.edu.sg\textsuperscript{1}, lee\_kong\_aik@i2r.a-star.edu.sg\textsuperscript{2}, ewsgan@ntu.edu.sg\textsuperscript{3}} \\}

\date{}

\begin{document}
\maketitle
\begin{abstract}
Conventional approaches to personalized dialogue generation typically require a large corpus, as well as predefined persona information. However, in a real-world setting, neither a large corpus of training data nor persona information are readily available. To address these practical limitations, we propose a novel multi-task meta-learning approach which involves training a model to adapt to new personas without relying on a large corpus, or on any predefined persona information. Instead, the model is tasked with generating personalized responses based on only the dialogue context. Unlike prior work, our approach leverages on the provided persona information only during training via the introduction of an auxiliary persona reconstruction task. In this paper, we introduce 2 frameworks that adopt the proposed multi-task meta-learning approach: the Multi-Task Meta-Learning (MTML) framework, and the Alternating Multi-Task Meta-Learning (AMTML) framework. Experimental results show that utilizing MTML and AMTML results in dialogue responses with greater persona consistency.  
\end{abstract}

\section{Introduction}

Personalized dialogue generation involves generating dialogue responses which incorporates the personality of the interlocutors, leading to more natural and human-like dialogue. Thus far, approaches to personalized dialogue generation typically require a large number of persona-specific dialogue examples. Certain approaches also require persona information presented in the form of several predefined persona statements (eg. 'I love dogs', 'I am an engineering student.'). However, in a real-world system, large amounts of persona-specific dialogue are rarely available, and collecting descriptive persona statements from every interlocutor is intractable.

To address these practical issues, Persona Agnostic Meta-Learning (PAML) \citep{madotto-etal-2019-personalizing}, a framework which aims to train a model capable of rapid adaptation to new unseen personas, was proposed. The PAML framework was based on the popular Model-Agnostic Meta-Learning (MAML) framework \citep{pmlr-v70-finn17a}. The recently proposed Customized Model Agnostic Meta-Learning (CMAML) \citep{song-etal-2020-learning} framework largely follows the PAML framework, with the exception of an additional network structure optimization component. Both the PAML and CMAML frameworks were benchmarked on the PersonaChat corpus \citep{zhang-etal-2018-personalizing}, a popular personalized dialogue generation corpus which provides persona statements describing each interlocutor in addition to the persona-specific dialogues. As it is unfeasible to collect persona statements from interlocutors in a real world setting, the PAML framework does not utilize the available persona statements during both meta-learning and inference. However, even though it is impractical to utilize the persona statements during inference, the persona statements can be used \emph{during meta-learning} to further improve model performance.

Hence, we introduce a novel multi-task meta-learning approach which leverages predefined persona statements \emph{only during meta-learning} via an additional persona reconstruction task. Essentially, this task involves generating all corresponding persona statements in its entirety given the dialogue context. We hypothesize that the introduction of the persona reconstruction task would result in parameters capable of effectively inducing the persona information from the dialogue context, which would lead to the generation of persona consistent dialogue. The persona statements are \emph{not} used during inference. Prior usage of multi-task learning for personalized dialogue generation involved the addition of a persona classification task \citep{9025776, Su2019PersonalizedDR} and a distraction utterance binary classification task \citep{icaart21}. To our knowledge, this is the first attempt at incorporating persona statement reconstruction.

Our contributions include 2 multi-task meta-learning frameworks which leverage the persona reconstruction task only during training. The Multi-Task Meta-Learning (MTML) framework, as well as a variant known as the Alternating Multi-Task Meta-Learning (AMTML) framework. While both MTML and AMTML involve the addition of a persona reconstruction task only during meta-learning, MTML involves combining the losses derived from generating the response and reconstructing the persona. AMTML, on the other hand, functions by constantly alternating between both tasks. Experimental results on the PersonaChat corpus reveal that utilizing MTML and AMTML result in responses which reflect the interlocutor’s persona to a larger extent compared to prior work.

\section{Methodology}

\begin{figure*}[htbp]
\centering
\subfloat[]{\includegraphics[height=2in,width=0.23\textwidth]{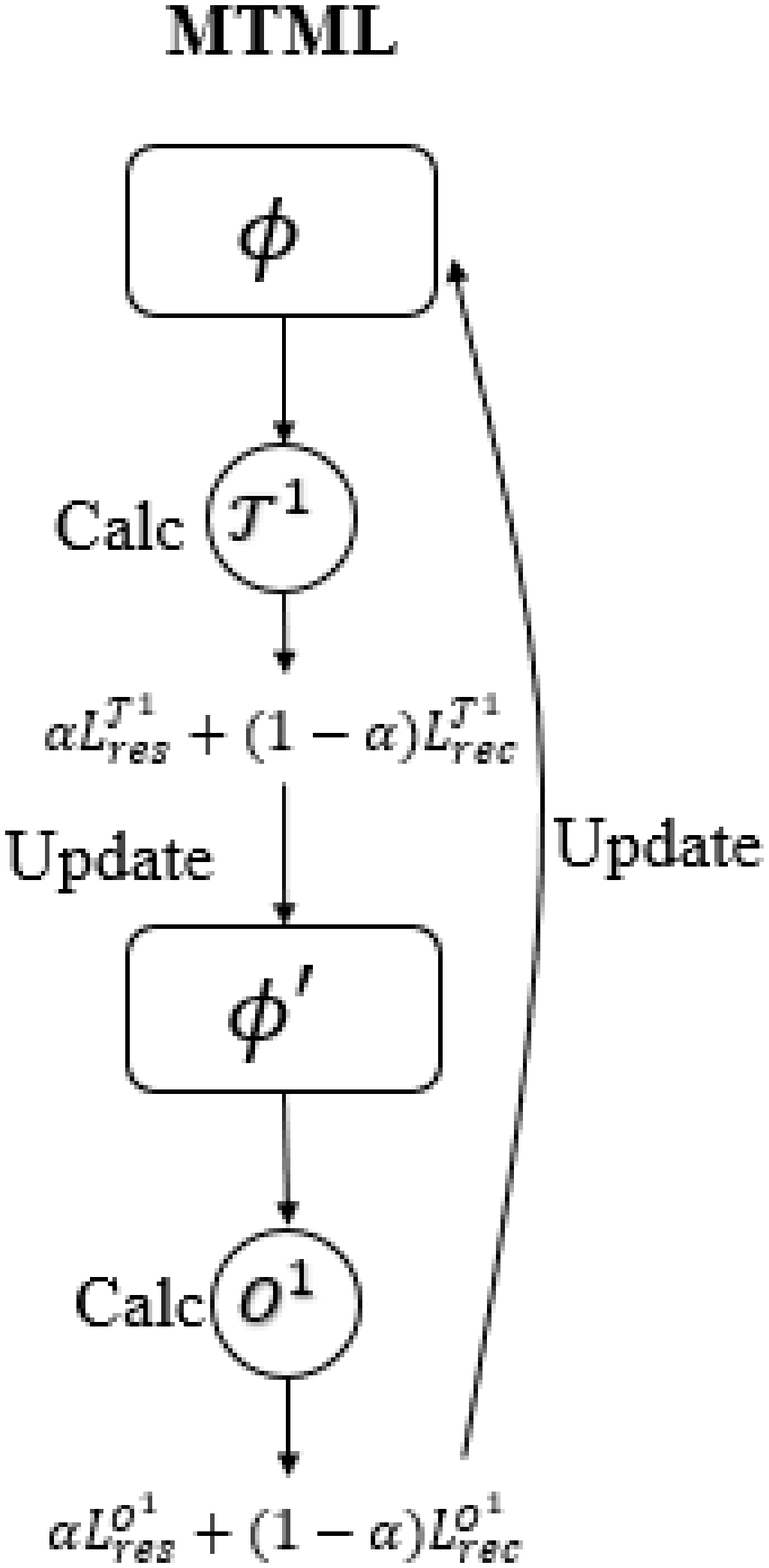}}
\qquad
\qquad
\qquad
\subfloat[]{\includegraphics[height=2in,width=0.3\textwidth]{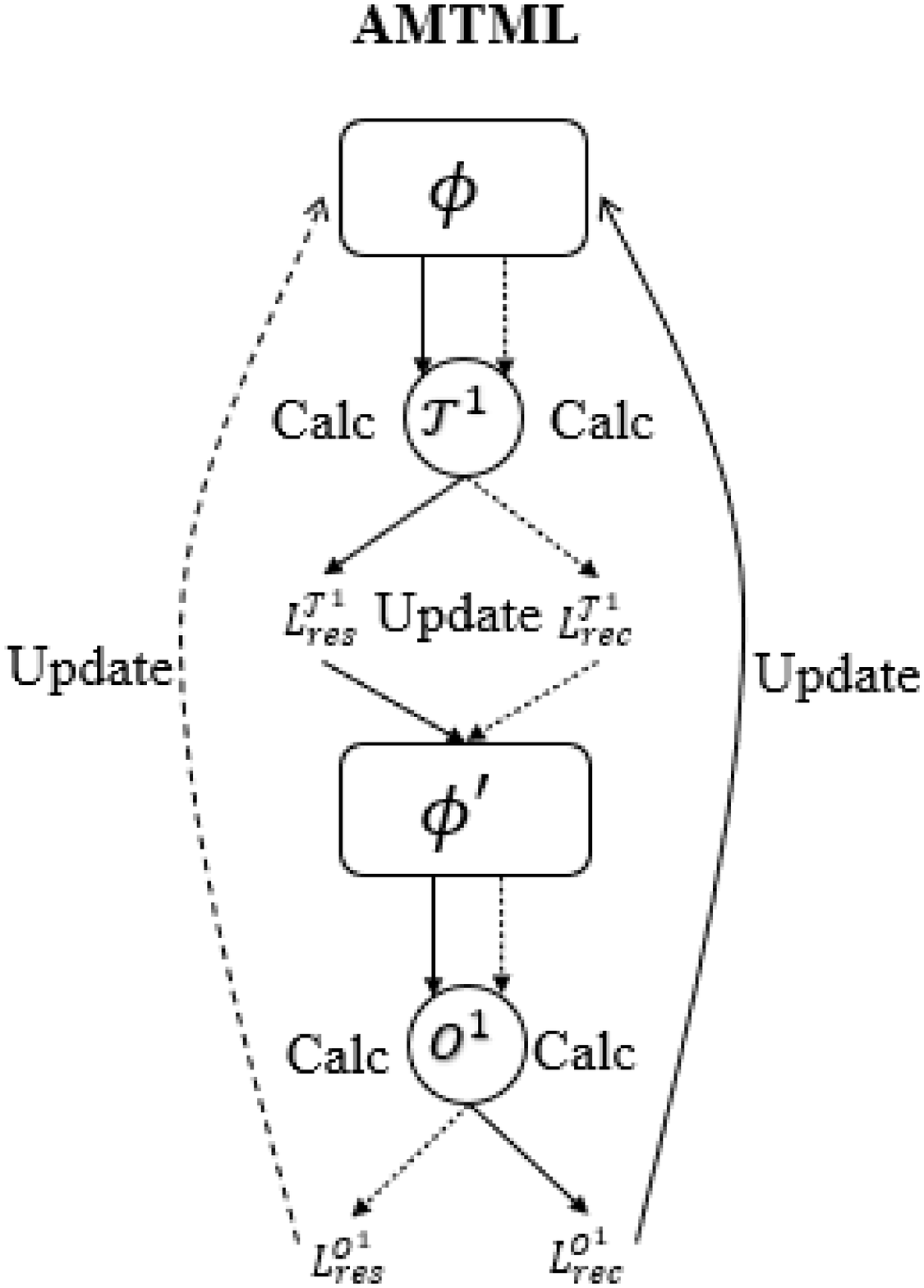}}
\caption{Diagrams depicting the computation path for MTML and AMTML. A batch size of 1 is assumed. (a): Diagram depicting the MTML framework. The weighted sum of the persona reconstruction loss and the response generation loss is used to update the parameters $\phi$. (b): Diagram depicting the AMTML framework. The computation flow alternates between the solid arrows and the dotted arrows after every iteration.}
\end{figure*}

\begin{table*}[]
\begin{tabular}{@{}ll@{}}
\toprule
\multicolumn{2}{c}{\textbf{Persona}}                \\ \midrule
\multicolumn{2}{l}{i have three guns and love hunting}             \\
\multicolumn{2}{l}{my family lives down the street from me}           \\
\multicolumn{2}{l}{i go to church every sunday} \\
\multicolumn{2}{l}{i drive a ford pickup truck}             \\
\multicolumn{2}{l}{i am very conservative}    \\ \midrule
\multicolumn{2}{c}{\textbf{Dialogue Context}}       \\ \midrule
User1:                       & hey there how are you i am shy                \\
User2:                       & well i am conservative so sounds like a match to me                     \\
User1:                       & i like that , did you go to college ?                     \\
User2:                       & no i just got my high school diploma and opened my owns shooting range                     \\
User1:                       & i went an got a computer science degree                      \\
User2:                       & wish i had continued school but i love what i do hunting , going to church and driving                     \\
User1:                       & gotcha , i build spaceships , the models                     \\
User2:                       &  o ok my family lives down from me and they build garages and farms                    \\
User1:                       & that is cool , my mom is a doctor                     \\ \midrule
\multicolumn{2}{c}{\textbf{Responses}}              \\ \midrule
Ref                          & you ever eat at wendys ? i went to washington school before i quit .                     \\
\multicolumn{2}{c}{\textbf{10-shot}}                \\
$Std$                          & that is cool where are you from ?                      \\
$Std_{p}$                         & that is cool what do you do for work ?                     \\
$PAML$                         & he is good to church every week                     \\
$MTML_{0.8}$                        & sure , i pick him up for church every sunday with my ford pickup\\
$AMTML$                        & my parents live in the house i grew up in , just down the street .                      \\
\multicolumn{2}{c}{\textbf{5-shot}}                 \\
$Std$                          & that is cool what kind of doctor do you do ?                     \\
$Std_{p}$                         & that is cool what kind of garages do you drive ?                     \\
$PAML$                         &  that is good where do you live ?                    \\
$MTML_{0.8}$                           &  that is good what do you do for work ?                     \\
$AMTML$                        & oh okay lol what do you do for work ?                      \\ \bottomrule
\end{tabular}
\caption{\label{tab:table-name} Dialogue responses generated by the implemented models.}
\end{table*}

Our approach to personalized dialogue generation involves extending the PAML framework \citep{madotto-etal-2019-personalizing} by introducing a persona reconstruction task only during the meta-learning stage. The PAML framework is essentially an adaptation of the general MAML framework for personalized dialogue generation. The MAML framework involves learning a parameter initialization capable of generalizing and adapting rapidly to new tasks unseen during the training process via gradient descent. Specifically, this involves obtaining the updated parameters by adjusting the model parameters by utilizing data from several related tasks. The original parameters are then optimized based on the updated parameters and a separate test set by computing the second order derivatives (Hessian matrix). In the PAML framework, each persona is viewed as a unique task. Consequently, the goal of the PAML framework is to obtain a parameter initialization capable of rapidly adapting to new personas. We hypothesize that the introduction of the persona reconstruction task would result in parameters which induce the persona from the dialogue context to a larger extent.

\subsection{Multi-task Learning}
 Unlike traditional approaches that involve utilizing both the dialogue context and persona information as model inputs during training, our framework involves reconstructing the persona statements as well as generating the dialogue response given the dialogue context. 
The primary task of generating the response and the corresponding loss function $L_{res}$ can be expressed in the following equations:
\begin{equation}
    f_{\phi}(x_{t}|x_{1:t-1}) = p(x_{t}|x_{1:t-1};\phi)
\end{equation}
\begin{equation}
    L_{res}(\phi) = - \sum_{t=1}^{T} log \: p(x_{t}|x_{1:t-1};\phi)
\end{equation}
where $x_{t}$ and $x_{1:t-1}$refer to the response and dialogue context respectively, and $\phi$ represents the model parameters. We hypothesize that the persona reconstruction task would result in model parameters capable of inducing the persona information from the dialogue context to a larger extent. This is due to the increased emphasis on persona consistency in the task of persona reconstruction.  Also, rather than generating only selected keywords or phrases, the persona reconstruction task involves generating all corresponding persona statements in its entirety given the dialogue context. This is because generating complete sentences would also require the model to account for fluency in addition to persona consistency, which is also a vital aspect of dialogue generation. The persona statements $\mathcal{P}_{1:N}$ are concatenated to form a single sequence $\overline{\mathcal{P}}$. The auxiliary persona reconstruction task and the corresponding loss function $L_{rec}$ is formalized in the following expression:
\begin{equation}
    \overline{\mathcal{P}} = concat(\mathcal{P}_{1:N})
\end{equation}
\begin{equation}
    f_{\phi}(\overline{\mathcal{P}}|x_{1:t-1}) = p(\overline{\mathcal{P}}|x_{1:t-1};\phi)
\end{equation}
\begin{equation}
    L_{rec}(\phi) = - \sum_{t=1}^{T} log \: p(\overline{\mathcal{P}}|x_{1:t-1};\phi)
\end{equation}

During training, the persona reconstruction loss and the response generation loss will be weighted and summed. Hence, the total multi-task loss is expressed as:
\begin{equation}
    L(\phi) = \alpha L_{res}(\phi) + (1 - \alpha) L_{rec}(\phi)
\end{equation}
where $\alpha$ determines the contribution of the persona reconstruction and response generation loss to the learning process.

\subsection{Multi-Task Meta Learning Framework}
Similar to PAML, MTML aims to learn a set of parameters $\phi$ capable of quickly adapting to an unseen persona in order to generate contextually appropriate responses which reflect the persona, without relying on user persona statements. However, unlike PAML, which does not utilize the persona statements during both meta-learning and inference, we leverage the persona information \emph{only during meta-learning}. The persona statements are not used during inference. The persona information is incorporated during meta-learning via the multi-task learning (Section 3.1). 

We begin by dividing the corpus into a training set, validation set and a test set denoted by $\mathcal{D}_{train}$, $\mathcal{D}_{valid}$ and $\mathcal{D}_{test}$ respectively. For each iteration $i$, a batch distinct of personas $\mathcal{P}_{1:N}^{1:M}$ and a corresponding set of $M$ dialogues are randomly sampled from the training set $\mathcal{D}_{train}$ to form the meta training set $\mathcal{T}^{i}$ (support set). Then, another set of $M$ dialogues corresponding to the same batch of personas $\mathcal{P}_{1:N}^{1:M}$ are randomly sampled from the training set $\mathcal{D}_{train}$ to form the meta optimization set $\mathcal{O}^{i}$ (query set). We refer to the meta training set and meta optimization set collectively as $\mathcal{C}^{i}$ i.e.,  $\mathcal{C}^{i}$ = ($\mathcal{T}^{i}$, $\mathcal{O}^{i}$).

During meta-training, the multi-task loss $L^{\mathcal{T}^{i}}_{mul}$ is computed via a weighted sum between the response generation loss $L^{\mathcal{T}^{i}}_{res}$ and the persona recreation loss $L^{\mathcal{T}^{i}}_{rec}$. To calculate $L^{\mathcal{T}^{i}}_{rec}$, the persona statements $\mathcal{P}_{1:N}^{1:M}$ of each persona were concatenated into a single sequence, resulting in $\overline{\mathcal{P}}^{1:M}$. For an arbitrary persona in the meta training set, the equations for computing the multi-task loss are as follows:
\begin{equation}
    L_{res}^{\mathcal{T}^{i}}(\phi) = - \sum log \: p(x_{t},|x_{1:t-1};\phi)
\end{equation}
\begin{equation}
    \overline{\mathcal{P}}^{1:M} = concat(\mathcal{P}_{1:N}^{1:M})
\end{equation}
\begin{equation}
    L_{rec}^{\mathcal{T}^{i}}(\phi) = - \sum log \: p(\overline{\mathcal{P}}|x_{1:t-1};\phi)
\end{equation}
\begin{equation}
    L^{\mathcal{T}^{i}}(\phi) = \alpha L_{res}^{\mathcal{T}^{i}}(\phi) + (1 - \alpha) L_{rec}^{\mathcal{T}^{i}}(\phi)
\end{equation}
where $\alpha$ accounts for the distribution between the 2 losses. Subsequently, the parameters $\phi$ are updated via SGD. The updated model parameters $\phi'$ can be expressed as:
\begin{equation}
    \phi' = \phi - \eta_{t} \nabla_{\phi} L^{\mathcal{T}^{i}}(\phi)
\end{equation}
where $\eta_{t}$ refers to the inner loop learning rate and $L^{\mathcal{T}^{i}}(\phi)$ represents the multi-task training loss attained. 

During meta-optimization, the original model $\phi$ is optimized on the multi-task loss attained on the meta optimization set $\mathcal{O}_{i}$ and the updated parameters $\phi'$. This meta-objective function can be expressed as:
\begin{equation}
\begin{aligned}
&\min_{\phi} \sum_{\mathcal{C}^{i} \sim \mathcal{D}_{train}} L^{\mathcal{O}^{i}} (\phi')\\
&= \sum_{\mathcal{C}^{i} \sim \mathcal{D}_{train}} L^{\mathcal{O}^{i}} (\phi - \eta_{t} \nabla_{\phi} L^{\mathcal{T}^{i}}(\phi))
\end{aligned}
\end{equation}
where $L^{\mathcal{O}^{i}}$ refers to the multi-task loss attained on the sampled meta optimization set $\mathcal{O}^{i}$. To obtain $L^{\mathcal{O}^{i}}(\phi')$, we first compute $L_{res}^{\mathcal{O}^{i}}(\phi')$ and $L_{rec}^{\mathcal{O}^{i}}(\phi')$ by applying Equation 7 - 9 on $\mathcal{O}^{i}$. Then, we compute the weighted sum between both losses:
\begin{equation}
    L^{\mathcal{O}^{i}}(\phi') = \alpha L_{res}^{\mathcal{O}^{i}}(\phi') + (1 - \alpha) L_{rec}^{\mathcal{O}^{i}}(\phi')
\end{equation}

where $\alpha$ determines the contribution of the persona reconstruction and response generation loss respectively. When $\alpha = 1$, MTML is analogous to MAML/PAML. Next, we sum the $L^{\mathcal{O}^{i}}(\phi')$ attained for every sampled persona $\mathcal{P}_{1:N}^{1:M}$. The original parameters $\phi$ are then updated using the average of losses obtained by dividing the summed loss by the batch size $M$. This can be formalized as:
\begin{equation}
    \phi = \phi - \eta_{o} \nabla_{\phi} \frac{1}{M} \sum_{\mathcal{O}^{i}} L^{\mathcal{O}^{i}}(\phi')
\end{equation}
where $\eta_{o}$ refers to the outer loop learning rate and $L^{\mathcal{O}^{i}}(\phi')$ represents the multi-task training loss attained by the updated parameters $\phi'$ on the sampled validation set $\mathcal{O}^{i}$. This computation involves obtaining the gradient of a gradient i.e., second order differentiation. A summary of MTML framework is provided in Algorithm 1. Additionally, an overview of the MTML framework is provided in Figure 1(a). 

\begin{algorithm}
    \caption{MTML}\label{your_label}
    \begin{algorithmic}
    \REQUIRE Hyperparameters $\alpha$, $\eta_{t}$, $\eta_{o}$
    \REQUIRE Dataset $\mathcal{D}_{train}$
        \FOR{iteration $i$ = $1$ to $n$}
        \STATE Sample $\mathcal{C}^{i} = (\mathcal{T}^{i},\mathcal{O}^{i}) \sim \mathcal{D}_{train}$
        \FOR{each persona in $\mathcal{T}^{i}$}
        \STATE Calc {$L^{\mathcal{T}^{i}}(\phi) = \alpha L_{res}^{\mathcal{T}^{i}}(\phi)+ (1 - \alpha) L_{rec}^{\mathcal{T}^{i}}(\phi)$}\STATE Update {$\phi'= \phi - \eta_{t} \nabla_{\phi} L^{\mathcal{T}^{i}}(\phi)$}
        \STATE Calc ${L^{\mathcal{O}^{i}}(\phi')=\alpha L_{res}^{\mathcal{O}^{i}}(\phi')+(1-\alpha) L_{rec}^{\mathcal{O}^{i}}(\phi')}$
        \ENDFOR
        \STATE Update $\phi = \phi - \eta_{o} \nabla_{\phi}  \frac{1}{M} \sum_{\mathcal{O}^{i}}L^{\mathcal{O}^{i}}(\phi')$
        \ENDFOR
    \end{algorithmic}
\end{algorithm}

\subsection{\textbf{Alternating Multi-Task Meta-Learning Framework}}

Instead of combining the response generation loss and persona reconstruction loss, Alternating-MTML(AMTML) involves constantly alternating between the two loss functions. Essentially, at every iteration, meta-training and meta-optimization are conducted with different loss functions. For AMTML, the $\alpha$ parameter would not be used. At every iteration, during meta-training, either the response generation loss $L_{res}$ or the persona reconstruction loss $L_{rec}$ will be used to update the parameters in the inner loop. Then, the alternate loss would be used to compute the loss for meta-optimization in the outer loop. For example, if the response generation loss is used to update the parameters during meta-training i.e. $L^{\mathcal{T}^{i}}_{res}$, the persona reconstruction loss would be used to derive the meta-optimization loss i.e. $L^{\mathcal{O}^{i}}_{rec}$. 

In our implementation, this is achieved by utilizing the response generation loss during meta-training when the iteration count is even, and utilizing the persona reconstruction loss when during meta-optimization when the iteration count is odd. Due to the alternating loss functions, the computational complexity and memory requirements for AMTML is lower than MTML, which requires computing both loss functions during both meta-training and meta-optimization. A summary of the AMTML framework is provided in Algorithm 2. Additionally, an overview of the AMTML framework is provided in Figure 1(b). The training set $\mathcal{D}_{train}$ is used to train the model and the validation set  $\mathcal{D}_{valid}$ is used to facilitate early stopping. 

\begin{algorithm}
    \caption{Alternating MTML}\label{your_label}
    \begin{algorithmic}
    \REQUIRE Hyperparameters $\eta_{t}$, $\eta_{o}$
    \REQUIRE Dataset $\mathcal{D}_{train}$
        \FOR{iteration $i$ = $1$ to $n$}
        \STATE Sample $\mathcal{C}^{i} = (\mathcal{T}^{i},\mathcal{O}^{i}) \sim \mathcal{D}_{train}$
        \FOR{each persona in $\mathcal{T}^{i}$}
        \IF{$i$ is even}
            \STATE Calc $L^{\mathcal{T}^{i}}_{res}(\phi)$
            \STATE Update $\phi'= \phi - \eta_{t} \nabla_{\phi} L^{\mathcal{T}^{i}}_{res}(\phi)$
            \STATE Calc $L^{\mathcal{O}^{i}}(\phi') = L_{rec}^{\mathcal{O}^{i}}(\phi')$
        \ELSIF{$i$ is odd}
            \STATE Calc $L_{rec}^{\mathcal{O}^{i}}(\phi')$
            \STATE Update $\phi'= \phi - \eta_{t} \nabla_{\phi} L^{\mathcal{T}^{i}}_{rec}(\phi)$
            \STATE Calc $L^{\mathcal{O}^{i}}(\phi') = L_{res}^{\mathcal{O}^{i}}(\phi')$
        \ENDIF
        \ENDFOR
        \STATE Update $\phi = \phi - \eta_{o} \nabla_{\phi}  \frac{1}{M} \sum_{\mathcal{O}^{i}}  L^{\mathcal{O}^{i}}(\phi')$
        \ENDFOR
    \end{algorithmic}
\end{algorithm}

\section{Experiment and Results}
\subsection{Corpus}
The proposed MTML and AMTML frameworks were evaluated on the PersonaChat dialogue corpus (Zhang et al., 2018). The corpus comprises 1155 distinct personas, each consisting of several persona statements. The PersonaChat dialogue corpus was chosen due to the available persona statements which are used to compute the persona reconstruction loss. In our experiment, the corpus is divided into a training, validation and test set. The validation and test sets each consist of 100 unique personas. For our experiments, we utilize the training and validation sets during the meta-learning stage and the test set during the testing stage.

\subsection{Implementation}
Following Madotto et al., we adopt the standard Transformer architecture \citep{NIPS2017_3f5ee243} consisting of 6 encoder layers, 6 decoder layers and 4 attention heads is used along with the GloVe embedding \citep{pennington-etal-2014-glove}. The dimensions of the word embedding and hidden dimension of the Transformer are fixed at 300. 
We use SGD ($\eta_{t}$ = 0.005, $M$ = 16) during meta-training and Adam ($\eta_{o}$ = 0.003, $M$ = 16) during meta-optimization. For MTML, we define an additional hyperparameter, $\alpha$, which accounts for the distribution between the persona recreation loss and the response generation loss.
\subsection{Evaluation}
\textbf{Automatic Metrics}
Similar to Madotto et al., we compute the BLEU score \citep{papineni-etal-2002-bleu}, the perplexity (PPL) and the C score \citep{madotto-etal-2019-personalizing} to evaluate the quality of the generated response. The BLEU score measures the similarity between the generated response and the reference response. PPL is the negative log of the generated response. The C-score reflects the amount of persona information present in the generated response by measuring the persona consistency  with respect to the corresponding persona statements via a BERT-based Natural Language Inference (NLI) model, which is finetuned to indicate if the generated response entails or contradicts the corresponding persona statements. A high C score would imply greater persona consistency. 

\noindent\textbf{Human Evaluation} We engaged 3 graduated individuals to evaluate 50 responses for each model using 3 criteria: persona consistency, fluency and contextual coherence.  Consistency reflects the amount of persona information corresponding to the persona statements are reflected in the generated response. Fluency accounts for any grammatical, spelling and phrasing issues, while Coherence reflects the appropriateness of the dialogue with respect to the dialogue history. Responses were assigned a rating from -1 to 1 for each criteria. For consistency, -1 indicates contradiction, 0 indicates neutrality and 1 indicates persona consistency (i.e. the response accurately reflects information from the corresponding persona statements). For coherence, the individuals were told to assign a score of -1 for incoherent, contextually illogical responses, 0 for a moderate coherent responses and 1 for coherent, contextual responses. Finally, for fluency, -1 is assigned to responses with several fluency issues, 0 for responses with one or two fluency errors and 1 for perfectly fluent responses.
\subsection{Experimental Settings}

We benchmark MTML/AMTML with the PersonaChat corpus. We train the following Transformer models:

$Std$: We pretrain a standard Transformer model with only the dialogue context as input. 

$Std_{p}$: We pretrain a standard Transformer model with both the dialogue context and persona statements as inputs. 

$PAML$: We pretrain a standard Transformer via PAML (Madotto et al., 2019). 

$MTML_{\alpha}$: We pretrain a standard Transformer trained via MTML (Section 3.2) where $\alpha$ = [0.9, 0.8, 0.7, 0.6, 0.5].
$AMTML$: We pretrain a standard Transformer via AMTML (Section 3.3).

During testing, the pretrained models described above were further trained, or finetuned, on dialogues corresponding to personas from the test set $\mathcal{D}_{test}$. As highlighted earlier, the models were finetuned using only the dialogue context, which was constructed by concatenating all previous utterances in the dialogue. For our experiment, the length of the dialogue context in each dialogue example would vary according to the number of turns. No restriction was placed on the number of dialogue turns required. No persona statements were used during finetuning. In the 5-shot and 10-shot setting, 5 and 10 dialogues corresponding to a persona was used to finetune the model parameters respectively. Then, the model tasked with generating the response corresponding to the same persona. Samples of the dialogue responses generated by each model is provided in Table 1. Table 2 and 3 depicts the results of automatic evaluation in a 10-shot and 5-shot setting respectively, while Table 4 and 5 depicts the results of the human evaluation in a 10-shot and 5-shot setting respectively.

\subsection{Results \& Discussion}

$MTML_{\alpha}$ generally achieves higher C-scores and Consistency scores compared to PAML, indicating responses which incorporate a larger amount of persona information. This confirms our hypothesis that the introduction of the persona reconstruction task during meta-learning would result in a model which induces the persona from the dialogue context to a larger extent.  However, it can be seen that PPL increases as $\alpha$ decreases. Since PPL scores have been found to correlate negatively with human likeness \citep{adiwardana2020humanlike}, a high PPL score is undesirable. This finding is supported by the human evaluation results, where the Fluency and Coherence scores drop as $\alpha$ increases in both the 5-shot and 10-shot settings. This implies that there is a trade-off between general fluency and persona consistency in the generated response. During meta-learning, if $\alpha$ is too large, the combined loss can be effectively reduced by minimizing the persona reconstruction loss. Hence, the model would be trained to generate responses which contain as much persona information as possible without considering fluency or context. While this would result in persona consistent responses, the responses would be largely incoherent and unnatural.Since $\alpha = 0.8$ strikes a balance between the PPL and C-scores, we conclude that the optimal value of $\alpha = 0.8$. 

\begin{table}[]
\centering
\begin{tabular}{|c|ccc|ccc|}
\hline
                        & PPL     & BLEU   & C-score     \\ \hline
$Std$          & 35.87         & 0.93        & 0.00          \\
$Std_{p}$          & 38.72        & 1.66       & 0.10           \\
$PAML$                 & 41.80         & 0.71        & 0.19 \\
$MTML_{0.5}$           & 77.32   & 0.53   & 0.46         \\
$MTML_{0.6}$           & 57.10   & 0.53   & 0.41          \\
$MTML_{0.7}$           & 52.44   & 0.57   & 0.47           \\
$MTML_{0.8}$           & 43.28   & 0.42   & 0.34          \\
$MTML_{0.9}$           & 40.39   & 0.71   & 0.21           \\
$AMTML$           & 48.66        & 0.48       & 0.29   \\ \hline
\end{tabular}
\caption{Automatic evaluation results (10-shot).}
\end{table}

\begin{table}[]
\centering
\begin{tabular}{|c|ccc|ccc|}
\hline
Method                       & PPL     & BLEU   & C-score  \\ \hline
$Std$          & 36.75         & 1.02       & -0.02          \\
$Std_{p}$          & 38.78     & 1.79       & 0.09           \\
$PAML$              & 40.46   & 0.65   & 0.15   \\
$MTML_{0.5}$        & 76.38   & 0.41   & 0.50    \\
$MTML_{0.6}$        & 55.19   & 0.53   & 0.48     \\
$MTML_{0.7}$        & 50.69   & 0.44   & 0.45     \\
$MTML_{0.8}$        & 41.42   & 0.38   & 0.30     \\
$MTML_{0.9}$        & 39.94   & 0.62   & 0.13     \\
$AMTML$        & 44.90   & 0.42   & 0.26   \\ \hline
\end{tabular}
\caption{Automatic evaluation results (5-shot).}
\end{table}

\begin{table}[]
\centering
\begin{tabular}{|c|ccc|ccc|}
\hline
                        & Consistency     & Fluency   & Coherence \\ \hline
$Std$                  & 0.16    & 0.92   & 0.24       \\
$Std_{p}$                  & 0.19    & 0.89   & 0.29       \\
$PAML$                 & 0.25    & 0.77   & 0.28      \\
$MTML_{0.5}$           & 0.47   & 0.24   & 0.20    \\
$MTML_{0.6}$           & 0.45   & 0.57   & 0.27      \\
$MTML_{0.7}$           & 0.45   & 0.60   & 0.30       \\
$MTML_{0.8}$           & 0.41 & 0.80    & 0.32     \\
$MTML_{0.9}$           & 0.39   & 0.88  & 0.17     \\
$AMTML$       & 0.42   & 0.89   & 0.33        \\ \hline
\end{tabular}
\caption{Human evaluation results (10-shot).}
\end{table}

\begin{table}[]
\centering
\begin{tabular}{|c|ccc|ccc|}
\hline
                        & Consistency     & Fluency   & Coherence \\ \hline
$Std$                  & 0.10    & 0.87   & 0.20       \\
$Std_{p}$                  & 0.09   & 0.89   & 0.21       \\
$PAML$                 & 0.13   & 0.84   & 0.27      \\
$MTML_{0.5}$           & 0.22   & 0.03   & -0.12    \\
$MTML_{0.6}$           & 0.24    & 0.65   & 0.15      \\
$MTML_{0.7}$           & 0.29   & 0.65   & 0.13       \\
$MTML_{0.8}$           & 0.22    & 0.78   & 0.29      \\
$MTML_{0.9}$           & 0.15    & 0.77  & 0.19     \\
$AMTML$           & 0.23   & 0.85   & 0.20        \\ \hline
\end{tabular}
\caption{Human evaluation results (5-shot).}
\end{table}

\begin{table}[]
\centering
\begin{tabular}{|l|ccc|l}
\cline{1-4}
Automatic & PPL         & BLEU    & C-score   &  \\ \cline{1-4}
$P^{2}Bot$     & \multicolumn{1}{c}{18.1}            & \multicolumn{1}{c}{0.61}        & \multicolumn{1}{c|}{0.33}          &  \\ \cline{1-4}
Human     & Consistency & Fluency & Coherence &  \\ \cline{1-4}
$P^{2}Bot$     & 0.39            & 0.91        &  0.43         &  \\ \cline{1-4}
\end{tabular}
\caption{Automatic and human evaluation results attained by $P^{2}Bot$}
\end{table}

Both $MTML_{0.8}$ and $AMTML$ improved the persona consistency of the generated responses. In terms of C-score, $MTML_{0.8}$ demonstrated a 78.9\%(10-shot) and 100\%(5-shot) improvement over $PAML$. In terms of persona consistency, $MTML_{0.8}$ demonstrated a 64.0\%(10-shot) and 15.4\%(5-shot) improvement over $PAML$. Similarly, compared to $PAML$, $AMTML$ achieved a 52.6\%(10-shot) and 73.3\%(5-shot) improvement when it comes to C-score. In terms of Consistency, $AMTML$ demonstrated a 68.0\%(10-shot) and 76.9\%(5-shot) improvement over $PAML$. Compared to $MTML_{0.8}$, in both the 10-shot and 5-shot settings, $AMTML$ achieved similar Consistency scores and slightly lower C-scores. However, while $MTML_{0.8}$ is comparable to $PAML$ with regard to Fluency and PPL, $AMTML$ outperformed $PAML$, $MTML_{0.8}$ and all other MTML variants in terms of Fluency. When it comes to coherence, $PAML$, $MTML_{0.8}$ and $AMTML$ generally achieved similar results.

Based on the results attained, while responses generated via MTML has a slight edge in terms of persona consistency, responses generated via AMTML are more fluent. On a side note, it should also be highlighted that the BLEU score did not correlate with any aspect of human evaluation. This further emphasizes the unsuitability of the BLEU score as an evaluation metric for dialogue generation \citep{liu-etal-2016-evaluate}. 

\subsubsection{PersonaChat SOTA Comparison}

Additionally, we compare our proposed frameworks with the current state-of-the-art framework for PersonaChat: $P^{2}Bot$ \citep{Song_Zhang_Hu_Liu_2020}. $P^{2}Bot$ involves finetuning the GPT pretrained language model on the training set via a transmitter receiver framework. This framework models the user's perception of the other party's persona in addition to the user's own persona. Hence, unlike MTML and AMTML, in the case of $P^{2}Bot$, persona statements are provided to the model along with the dialogue context during inference and testing. 

From Table 6, it can be observed that the C-score attained by $MTML_{0.8}$, in the 10-shot setting, was comparable to $P^{2}Bot$. When it comes to the Consistency score, in the 10-shot setting, both $MTML_{0.8}$ and $AMTML$ outperformed $P^{2}Bot$. This implies that the responses generated by $MTML_{0.8}$ and $AMTML$ generally reflect the corresponding persona information to a greater extent compared to $P^2Bot$ \emph{despite not being provided the persona statements during testing}. However, in terms of Fluency and Coherence, $P^{2}Bot$ still outperforms both $MTML_{0.8}$ and $AMTML$. This could be partially attributed to the use of the GPT pretrained model, which enhanced the overall quality of the generated responses.

\subsubsection{Persona Reconstruction}

In this section, we will provide a brief discussion regarding the persona reconstruction task. Based on the results attained, it is clear that the introduction of the persona reconstruction task during meta-learning further incentivizes the model to incorporate more persona information in the generated responses. Under the proposed frameworks, it is challenging to evaluate the performance of the model solely on the persona reconstruction task. However, based on the observed responses and loss values, persona reconstruction tend to be more successful when a longer dialogue context $x_{1:t-1}$ (greater number of turns) is provided. This is expected as a short dialogue context would not contain sufficient persona information for the model to reconstruct. 

Persona reconstruction is a interesting and challenging task that could be explored in future work. For this task, the model has to successfully infer the persona from the dialogue context as well as ensure the fluency of the generated description. Finetuning pretrained language models would be a good starting point for future work. Also, to prevent a mix-up between the personas from each interlocutor, only the dialogue utterances of the corresponding interlocutor should be utilized during training and inference.

\section{Related Work}

\textbf{Multi-task Learning} Multi-task learning broadly refers to the process of learning more than one tasks/objectives concurrently with a shared model. In addition to personalized dialogue generation, multi-task learning has been applied to task-oriented dialogue subtasks including response generation\citep{zhu-etal-2019-multi}, dialogue state tracking\citep{inproceedings, trinh-etal-2018-multi, rastogi-etal-2018-multi}, dialogue act selection \citep{mcleod-etal-2019-multi}, as well as conditional open-domain dialogue generation\citep{zeng-nie-2021-simple, DBLP:journals/corr/abs-2105-11696}.

\noindent{\textbf{Meta-learning}}
Meta-learning involves teaching models how to learn efficiently and quickly. There are 3 broad categories of meta-learning algorithms: optimization-based \citep{pmlr-v70-finn17a, Nichol2018OnFM}, metric-based \citep{NIPS2017_cb8da676, NIPS2016_90e13578}, and model-based \citep{Mishra2018ASN, pmlr-v48-santoro16}. Optimization-based meta-learning approaches, which involve directly updating the model’s parameters to allow for rapid adaptation to unseen tasks, have been applied to various dialogue tasks. Examples of such applications include task-oriented dialogue generation \citep{ijcai2019-437,dai-etal-2020-learning, peng-etal-2020-shot}, domain adaptation \citep{qian-yu-2019-domain} and dialogue state tracking \citep{peng-etal-2020-shot, huang-etal-2020-meta}. 

\noindent{\textbf{Personalized Dialogue Generation}}
There are numerous forms of personalized dialogue generation. The form covered in this paper requires leveraging both the persona information and dialogue context.  Another form of personalized dialogue generation involves conditioning the response on external profile/identity information. Thus far, many different architectures \citep{wu2020guiding,song2019exploiting,DBLP:journals/corr/abs-1901-08149, ijcai2017-521, DBLP:journals/corr/JoshiMF17} and training frameworks \citep{Song_Zhang_Hu_Liu_2020, liu-etal-2020-impress, DBLP:journals/corr/abs-1901-09672, Su2019} which involve utilizing the encoded persona/ or personality information and dialogue context as input have been proposed. For certain dialogue corpora such as DailyDialog \citep{li-etal-2017-dailydialog} and PERSONALDIALOG \citep{Zheng_Zhang_Huang_Mao_2020}, the persona descriptions are not provided. Instead, a representation of the interlocutor's personality should be inferred from the dialogue history.

\section{Conclusion}
In this work, we proposed MTML and AMTML, 2 meta-learning frameworks which adopt our multi-task learning approach involving the addition of a persona reconstruction task. Empirical results demonstrate that both MTML and AMTML effectively increases the amount of persona information reflected in the generated dialogue responses compared to prior work. However, there is still room for improvement when it comes to the fluency and contextual coherence of the generated responses. Future work could involve improving these aspects of the responses by incorporating pretrained language models in meta-learning framework.

\bibliographystyle{acl_natbib}
\bibliography{anthology,acl2021}

\end{document}